\documentclass[10pt,twocolumn,letterpaper]{article}

\usepackage{iccv}              

\definecolor{iccvblue}{rgb}{0.21,0.49,0.74}
\usepackage[pagebackref,breaklinks,colorlinks,allcolors=iccvblue]{hyperref}

\usepackage{booktabs}  
\usepackage{pifont}    
\usepackage{caption}   
\usepackage{array}     
\usepackage{multirow}  
\usepackage{colortbl}
\usepackage{xcolor}
\definecolor{cvprblue}{RGB}{0, 82, 147}
\usepackage{arydshln}
\usepackage{graphicx}
\usepackage{lipsum}
\usepackage{indentfirst}
\usepackage{amsmath}
\usepackage{cases}
\usepackage{booktabs}
\usepackage{multirow} 
\usepackage{graphicx}
\usepackage{times}  
\usepackage{algorithm}
\usepackage{algorithmic}
\usepackage{color}
\usepackage{ulem}
\usepackage{pifont}
\usepackage{arydshln} 
\usepackage{float} 
\usepackage{bm}
\usepackage{caption}
\usepackage{listings}
\def\paperID{***} 
\def\confName{CVPR}
\def\confYear{2025}

\title{EasyControl: Adding Efficient and Flexible Control for Diffusion Transformer}

\author{
Yuxuan Zhang\textsuperscript{1},
Yirui Yuan\textsuperscript{2,1},
Yiren Song\textsuperscript{3,1},
Haofan Wang\textsuperscript{4},
Jiaming Liu\textsuperscript{ 1}$^{\dagger}$ \\
\textsuperscript{1} Tiamat AI, 
\textsuperscript{2}ShanghaiTech University, \\
\textsuperscript{3}National University of Singapore,
\textsuperscript{4}Liblib AI \\
\url{https://github.com/Xiaojiu-z/EasyControl}
}

\begin{document}

\twocolumn[{
\maketitle
\begin{figure}[H]
\hsize=\textwidth 
\centering
\vspace{-1cm} 
\includegraphics[width=1\textwidth]{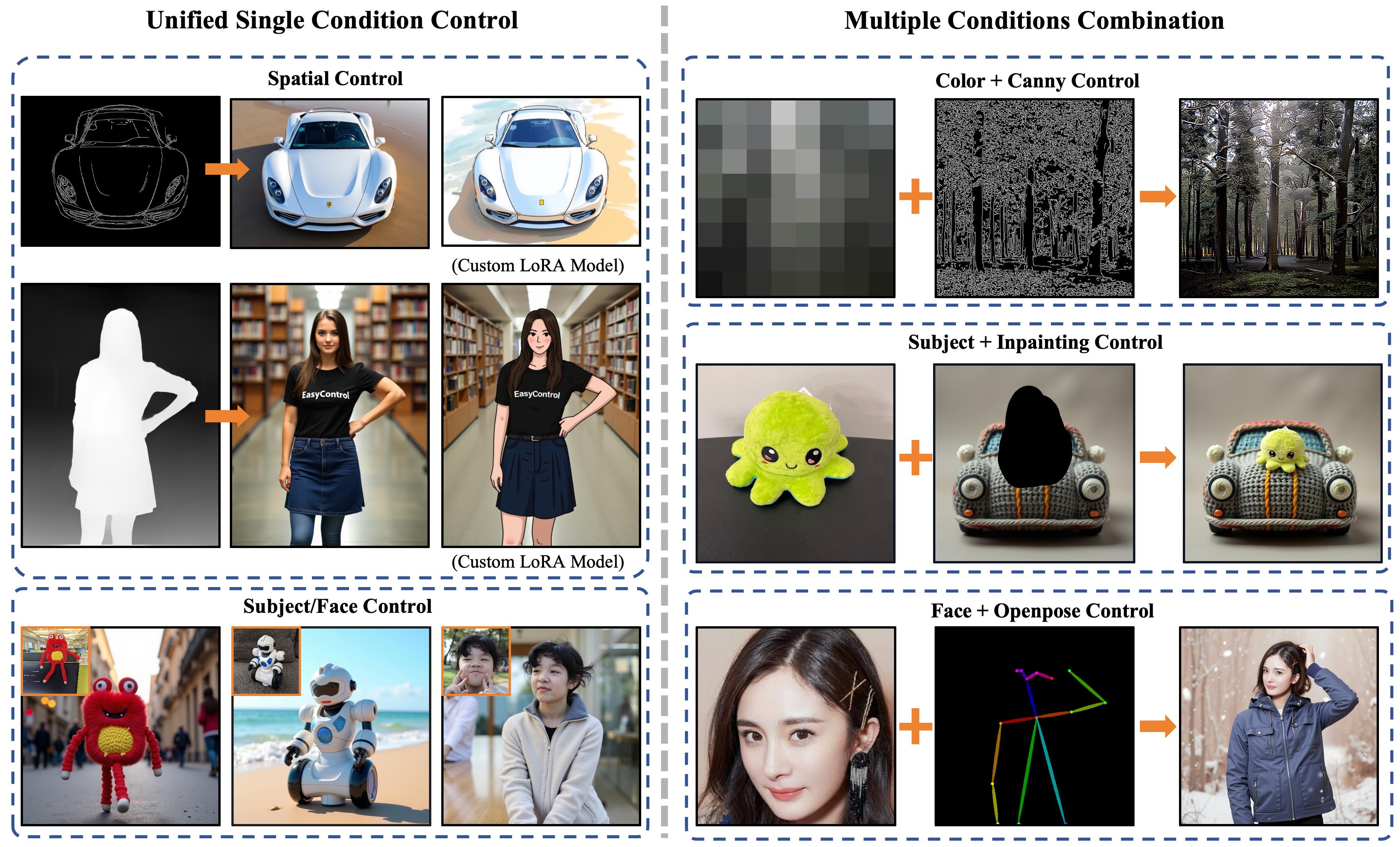}
\vspace{-0.7cm}

\caption{Our proposed framework, \textit{EasyControl}, is a lightweight and efficient plug-and-play module specifically designed for diffusion transformer. This solution not only enables spatial control and subject/face control under single conditions but also demonstrates remarkable zero-shot multi-condition generalization capability after single-condition training, achieving sophisticated multi-condition control.}
    \label{teaser}
\end{figure}
}]

\begin{abstract}
Recent advancements in Unet-based diffusion models, such as ControlNet and IP-Adapter, have introduced effective spatial and subject control mechanisms. However, the DiT (Diffusion Transformer) architecture still struggles with efficient and flexible control. To tackle this issue, we propose EasyControl, a novel framework designed to unify condition-guided diffusion transformers with high efficiency and flexibility. Our framework is built on three key innovations. First, we introduce a lightweight Condition Injection LoRA Module. This module processes conditional signals in isolation, acting as a plug-and-play solution. It avoids modifying the base model’s weights, ensuring compatibility with customized models and enabling the flexible injection of diverse conditions. Notably, this module also supports harmonious and robust zero-shot multi-condition generalization, even when trained only on single-condition data. Second, we propose a Position-Aware Training Paradigm. This approach standardizes input conditions to fixed resolutions, allowing the generation of images with arbitrary aspect ratios and flexible resolutions. At the same time, it optimizes computational efficiency, making the framework more practical for real-world applications. Third, we develop a Causal Attention Mechanism combined with the KV Cache technique, adapted for conditional generation tasks. This innovation significantly reduces the latency of image synthesis, improving the overall efficiency of the framework. Through extensive experiments, we demonstrate that EasyControl achieves exceptional performance across various application scenarios. These innovations collectively make our framework highly efficient, flexible, and suitable for a wide range of tasks.
\end{abstract}    
\section{Introduction}
In recent years, image generation systems based on diffusion models have undergone significant architectural evolution. The technical trajectory has gradually shifted from the early UNet-based architecture\cite{ldm,sdxl} to the Transformer-based DiT (Diffusion Transformer) model\cite{dit,flux2024,sd3}. During the UNet-based era, pre-trained models such as SD 1.5/XL established a thriving ecosystem, giving rise to a series of plug-and-play conditional generation extension modules, such as ControlNet\cite{controlnet}, IP-Adapter\cite{ipadapter}, and so on\cite{t2i-adapter,controlnet++,zhao2024uni}. These modules achieved flexible expansion of pre-trained models by freezing pre-trained parameters and introducing additional adapters or Encoder architectures, thereby promoting the widespread application of text-to-image generation technology. The DiT architecture's rise catalyzed a paradigm shift in the conditional generation, marking a transition toward token-based approaches in the field. These methods typically convert conditional images into image token sequences through a VAE encoder, concatenate them with noise latent representations, and finetune model\cite{lora} to achieve conditional-guided generation. This approach has demonstrated significant potential in applications such as virtual try-on\cite{any2anytryon,incontextlora,ITVTON}, image editing\cite{huang2025photodoodle,incontextlora,xia2024dreamomni,xiao2024omnigen,chen2024unireal,han2024ace,xie2025anyrefill}, spatial control generation\cite{tan2024ominicontrol,chen2024unireal,han2024ace,zhang2025eligen,xiao2024omnigen,xie2025anyrefill,xia2024dreamomni}, and subject-driven generation\cite{tan2024ominicontrol,chen2024unireal,han2024ace,incontextlora,xie2025anyrefill,xia2024dreamomni}.

However, despite notable advancements in existing methods, the DiT-based conditional generation systems still face several critical challenges compared to the mature UNet-based ecosystem. First, there is the issue of computational efficiency bottlenecks. When additional image tokens are introduced, the self-attention mechanism incurs quadratic time complexity relative to the input length due to its attention mechanism. Specifically, the model parameter count is proportional to the square of the number of tokens, which significantly increases inference latency and limits the expansion of practical application scenarios. Second, there is the challenge of multi-condition collaborative control. Existing methods struggle to achieve stable coordination under multi-condition guidance within a single-condition training paradigm. The representational conflicts of different conditional signals in the latent space lead to a degradation in generation quality, particularly in zero-shot multi-condition combination scenarios, where the model lacks effective cross-condition interaction mechanisms. Finally, there are limitations in model adaptability. Although current Parameter-Efficient Fine-Tuning methods\cite{lora} can keep the backbone network parameters frozen, there is parameter conflict between the fine-tuning modules and the custom models in the community. This design flaw results in feature degradation during style transfer, limiting the plug-and-play characteristics of the modules. Therefore, our goal is to construct a novel DiT conditional generation framework, providing the community with a DiT extension solution that combines efficient inference, flexible control, and plug-and-play features, thereby promoting the smooth migration of the diffusion model ecosystem to more efficient Transformer architectures.

This paper presents EasyControl, an efficient and flexible condition-guided DiT framework that achieves significant improvements in efficiency and flexibility through innovations at three fundamental levels.

At the model architecture level, we propose a lightweight and plug-and-play Condition Injection LoRA Module, whose core innovation lies in the isolated injection of condition signals. This module is integrated into pre-trained models through a parallel branch mechanism, where low-rank projection is exclusively applied to condition branch tokens while keeping the weights of text and noise branches frozen. This design ensures seamless compatibility with customized models while supporting harmonious and robust zero-shot multi-condition generalization, significantly expanding its applicability across diverse visual generation tasks.

At the token processing level, we introduce a novel Position-Aware Training Paradigm that enhances resolution flexibility and computational efficiency through two key innovations: (1) resolution normalization of control conditions to reduce input sequence length, and (2) Position-Aware Interpolation to maintain spatial consistency between condition tokens and noise tokens. This mechanism enables the model to effectively learn arbitrary aspect ratios and multi-resolution representations, even under low-resolution control signals.

At the attention mechanism level, we implement a transformation by replacing the conventional full attention with Causal Attention, integrated with the KV Cache technique. This represents the first implementation of a condition feature caching strategy based on the KV Cache mechanism. Specifically, at the initial diffusion time step (t = 0), the system precomputes and persistently stores key-value pairs of all condition features, which are then reused throughout subsequent time steps (t$\geq$1), achieving substantial computational savings.

Through these three-level innovations, EasyControl establishes a new paradigm in conditional generation, demonstrating superior efficiency and flexibility. In summary, our contributions are as follows:
\begin{itemize}
\item We EasyControl, a new paradigm in conditional generation for DiT models. In EasyControl, each condition undergoes an isolated condition branch, which is adapted from the pre-trained DiT model via the Condition Injection LoRA module. This design enables seamless integration with custom models. This approach supports flexible condition injection and effectively fuses multiple conditions.

\item \textbf{Efficiency:} Our framework achieves high efficiency through two key innovations. First, the Position-Aware Training Paradigm standardizes input conditions to fixed resolutions, ensuring adaptability and computational efficiency. Second, we integrate the Causal Attention Mechanism with the KV Cache technique, marking the first successful application of KV Cache in conditional generation tasks. This combination significantly reduces latency and boosts overall efficiency.

\item \textbf{Flexiblity:} EasyControl offers exceptional flexibility by enabling the generation of images with varying resolutions and aspect ratios. This design ensures robust performance across diverse scenarios, balancing high-quality generation with adaptability to different requirements.

\end{itemize}
\section{Related Works}
\label{sec:related}
\subsection{Diffusion Models}
Diffusion models have emerged as a powerful class of generative models, evolving from their initial formulation in denoising probabilistic frameworks to state-of-the-art architectures like DDPM\cite{ddpm}, DDIM\cite{ddim}, and LDM\cite{ldm}. These models have demonstrated remarkable success across a wide range of generative tasks, including text-to-image synthesis\cite{sdxl, sd3, ldm, flux2024,saharia2022photorealistic, ramesh2021zero}, text-to-video generation\cite{stablevideo,guo2023animatediff,hunyuanvideo,hong2022cogvideo,yang2024cogvideox,makeavideo,wan2024grid, song2025layertracer, song2025makeanything, song2024processpainter}, text-to-3D creation\cite{shi2023mvdream,poole2022dreamfusion,chen2023fantasia3d,lin2023magic3d,liu2024one,qian2023magic123}, and image editing\cite{huang2025photodoodle,shi2024seededit,attend,zeroimagetrans,stablehair,stablemakeup,wang2024stablegarment,nulltextinversion,brooks2023instructpix2pix,cao2023masactrl,kawar2023imagic,zhang2023magicbrush,li2024zone}. By iteratively refining noise into structured outputs, diffusion models have achieved unprecedented fidelity and diversity in generated content. Recently, the architecture of diffusion models has shifted from the traditional U-Net-based design to transformer-based frameworks, marking a significant milestone in the field. Models like SD3\cite{sd3} and FLUX\cite{flux2024} have adopted DiT\cite{dit} structures which have enabled improved image quality, higher resolution synthesis, and better handling of complex conditional inputs.

\subsection{Conditional Image Generation}
Condition-guided diffusion models aim to enhance the generation process by incorporating additional conditions, enabling precise control over the output. These models can be broadly categorized into two types: pixel-aligned and non-pixel-aligned methods. Pixel-aligned approaches, such as ControlNet\cite{controlnet} and so on\cite{t2i-adapter,tan2024ominicontrol,mo2024freecontrol,controlnet++,zhao2024uni,song2024idprotector, song2024anti}, directly align conditional inputs (e.g., edges, depth maps) with the generated output at the pixel level, ensuring fine-grained spatial control. On the other hand, non-pixel-aligned methods, often referred to as subject generation, focus on high-level semantic control. Representative works include IP-Adapter\cite{ipadapter}, SSR-Encoder\cite{ssr}, and so on\cite{Blip-diffusion,customdiff,e4t,xiao2024fastcomposer,subjectdiff,ruiz2023dreambooth,realcustom,MS-Diffusion,mipadapter,TI,fast_icassp}, which leverage adapters or encoders to inject subject-specific conditions into the diffusion process. 

Recent advancements in DiT-based conditional generation have demonstrated significant progress. These approaches primarily encode visual conditions into tokens and integrate them with text tokens. For instance, methods\cite{xiao2024omnigen, chen2024unireal, xia2024dreamomni, onediffusion} trained a DiT-based foundation model that tokenizes conditional images and jointly trains them with text tokens for diverse conditional visual tasks within a single framework. Methods like In-context LoRA\cite{incontextlora}, OminiControl\cite{tan2024ominicontrol}, and so on\cite{huang2025photodoodle,any2anytryon,xie2025anyrefill} introduce a novel approach by concatenating images or tokens and applying task-specific LoRA within pre-trained DiT tuning with task-specific datasets for various applications. While these methods represent notable progress in unifying control mechanisms, they exhibit several critical limitations. (1) Token concatenation incurs high computational costs. (2) Conventional LoRA designs struggle to handle flexible combinations of multiple conditions effectively. (3) Most methods lack open-source or plug-and-play support, hindering community adoption. Our work addresses these limitations by proposing a unified DiT-based framework that ensures efficiency and flexibility.
\section{Method}

\begin{figure*}[h]
    \centering
    \vspace{-0.6cm}
    \includegraphics[width=1.\linewidth]{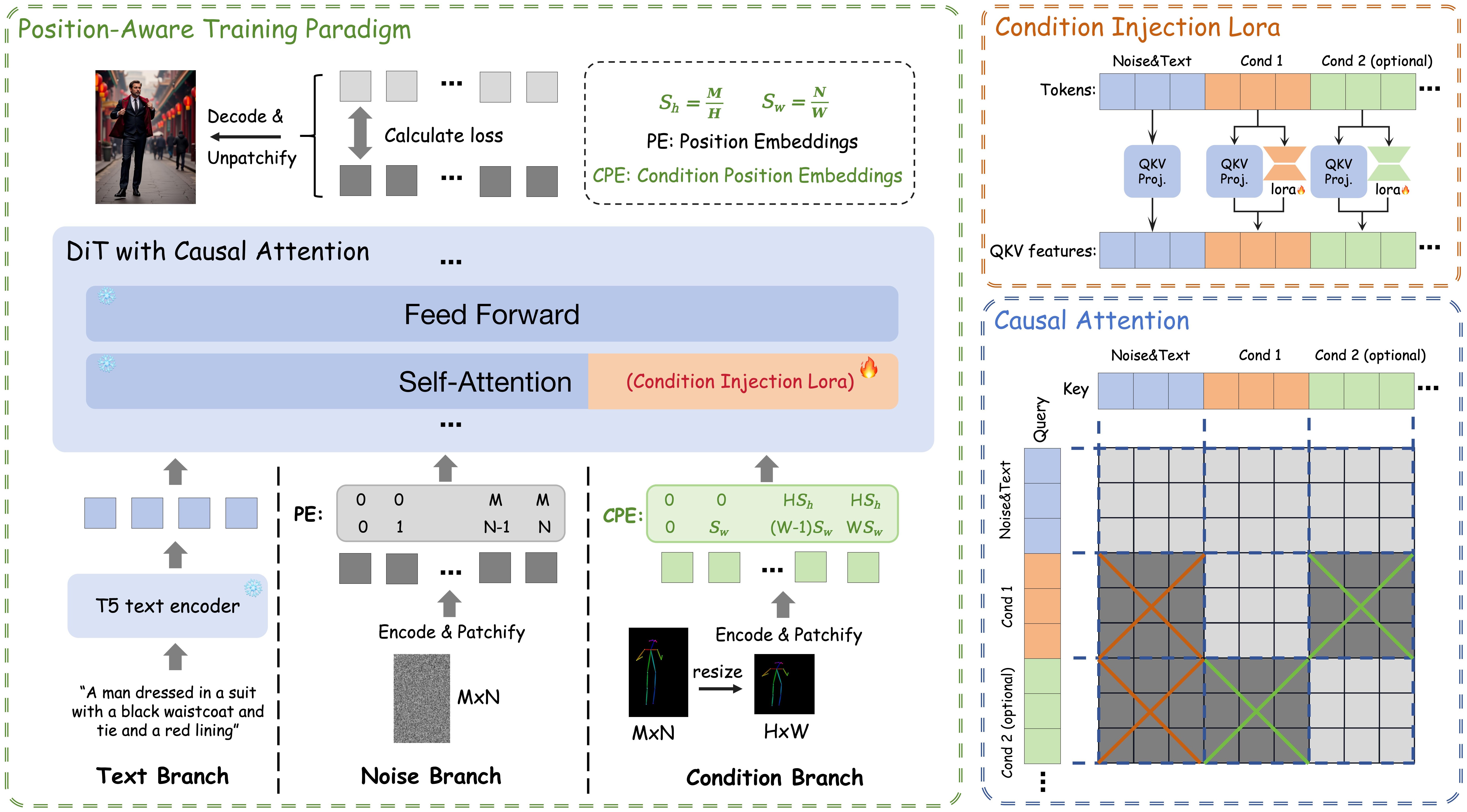}
    \caption{The illustration of EasyControl framework. The condition signal is injected into the Diffusion Transformer (DiT) through a newly introduced condition branch, which encodes the condition tokens in conjunction with a lightweight, plug-and-play Condition Injection LoRA Module. During training, each individual condition is trained separately, where condition images are resized to a lower resolution and trained using our proposed Position-Aware Training Paradigm. This approach enables efficient and flexible resolution training. The framework incorporates a Causal Attention mechanism, which enables the implementation of a Key-Value (KV) Cache to substantially improve inference efficiency. Furthermore, our design facilitates the seamless integration of multiple Condition Injection LoRA Modules, enabling robust and harmonious multi-condition generation.}
    \vspace{-0.5cm}
    \label{fig:method}
\end{figure*}

In this section, we present the technical details of EasyControl, and the overall framework of the method is illustrated in figure~\ref{fig:method}. EasyControl is built on a FLUX.1 dev. It comprises several key components: the Condition Injection LoRA Module (Sec.~\ref{clm}), the Causal Attention in EasyControl (Sec.~\ref{cca}), the Position-Aware Training Paradigm (Sec.~\ref{patp}), and the KV Cache for Inference (Sec.~\ref{kvcache}). The preliminaries on DiT are detailed in supplementary (Sec.A).


\subsection{Condition Injection LoRA Module}
\label{clm}
To efficiently incorporate conditional signals while preserving the pre-trained model’s generalization ability, we extend the FLUX architecture by introducing an additional \textbf{Condition Branch}. Unlike conventional approaches that introduce separate control modules, our method seamlessly integrates conditional information into the existing architecture while avoiding redundant parameters and computational overhead.

In Transformer-based architectures, input feature representations are first projected into \textit{query} (\( Q \)), \textit{key} (\( K \)), and \textit{value} (\( V \)) features before being processed by the self-attention mechanism. Given input representations \( Z_t, Z_n, Z_c \) corresponding to the \textbf{Denoising} (\textbf{Text}, \textbf{Noise}), and \textbf{Condition Branches}, the standard QKV transformation is defined as follows:
\begin{equation}
Q_i, K_i, V_i = W_Q Z_i, W_K Z_i, W_V Z_i, \quad i \in \{t, n, c\}
\end{equation}
where \( W_Q, W_K, W_V \) are the shared projection matrices across all branches. While this design allows for efficient parameter sharing, it does not explicitly optimize the representation of conditional signals. To address this limitation, we introduce \textit{LoRA (Low-Rank Adaptation)} to adaptively enhance the conditional representations while keeping other branches unmodified:
\begin{equation}
\Delta Q_c, \Delta K_c, \Delta V_c = B_Q A_Q Z_c, B_K A_K Z_c, B_V A_V Z_c
\end{equation}

Thus, the updated QKV features for the \textbf{Condition Branch} are:
\begin{equation}
Q'_c, K'_c, V'_c = Q_c + \Delta Q_c, K_c + \Delta K_c, V_c + \Delta V_c
\end{equation}
where \( A_i, B_i \in \mathbb{R}^{d \times r} \) (with \( r \ll d \), \( i \in \{Q, K, V\} \)) are low-rank matrices that parameterize the LoRA transformation.  
Notably, the \textit{Text} and \textit{Noise} branches remain unchanged:
\begin{equation}
Q'_i = Q_i, \quad K'_i = K_i, \quad V'_i = V_i, \quad i \in \{t, n\}
\end{equation}
By applying LoRA-based adaptation exclusively to the Condition Branch, we ensure that conditional signals are efficiently injected into the model without disrupting the pre-trained text and noise representations. This targeted adaptation allows the model to flexibly integrate conditional information while maintaining the integrity of its original feature space, thereby achieving more controllable and high-fidelity image generation.

\subsection{Causal Attention in EasyControl}
\label{cca}
Causal Attention is a unidirectional attention mechanism designed to restrict information flow in sequence models by allowing each position to attend only to previous positions and itself, thereby enforcing temporal causality. This is achieved by applying a mask with values of $0$ and $-\infty$ to the attention logits before the Softmax operation, mathematically expressed as:
\begin{equation}
\mathbf{Q} = [Q'_t; Q'_n; Q'_c],
\mathbf{K} = [K'_t; K'_n; K'_c],
\mathbf{V} = [V'_t; V'_n; V'_c]
\end{equation}
\begin{equation}
\textit{Softmax}(\mathbf{Q}\mathbf{K}^\top / \sqrt{d_k} + M) \mathbf{V}
\end{equation}
where $M$ enforces the causality constraint and QKV are the concatenated features from the noise, text, and condition branches. To improve the efficiency of inference and effectively integrate multiple conditional signals, we design two specialized causal attention mechanisms: Causal Conditional Attention and Causal Mutual Attention. These mechanisms govern information flow through distinct masking strategies to balance conditional aggregation and isolation.

\subsubsection{Causal Conditional Attention}
This mechanism enforces two rules: $(1)$ intra-condition computation within each condition branch and  $(2)$ an attention mask that prevents condition tokens from querying denoising $(\text{text\&noise})$ tokens during training. Formally, we define the input sequence in single-condition training as:
\begin{equation}
Z = [Z_{\text{t\&n}}; Z_{\text{c}}]
\end{equation}
where \( Z_{\text{t\&n}}\) denotes the noise and text tokens, and \( Z_{\text{c}} \) represents the condition tokens, we define an attention mask \( M \in \{0, -\infty\}^{n \times n} \) to regulate attention flow. Specifically, the mask is formulated as:
\begin{equation}
M_{ij} =
\begin{cases}
-\infty, & \text{if } i \notin n_{\text{t\&n}} \text{and} j\in n_{\text{t\&n}}  \\
0, & \text{otherwise}
\end{cases}
\end{equation}  
where \( n = n_{\text{t\&n}} + n_{\text{c}} \) represents the total sequence length. 

This design blocks unidirectional attention from the \textbf{condition branch} to the $\textbf{denoising \text{(noise\&text)} branch}$ while allowing denoising branch tokens to freely aggregate conditional signals. By strictly isolating condition-to-denoising queries, the design enables decoupled KV Cache states for each branch during inference, thereby reducing redundant computation and significantly improving image generation efficiency.

\subsubsection{Causal Mutual Attention}
\begin{figure}[!t]
\centering
    \vspace{-0.5cm}
\includegraphics[width=1.\linewidth]{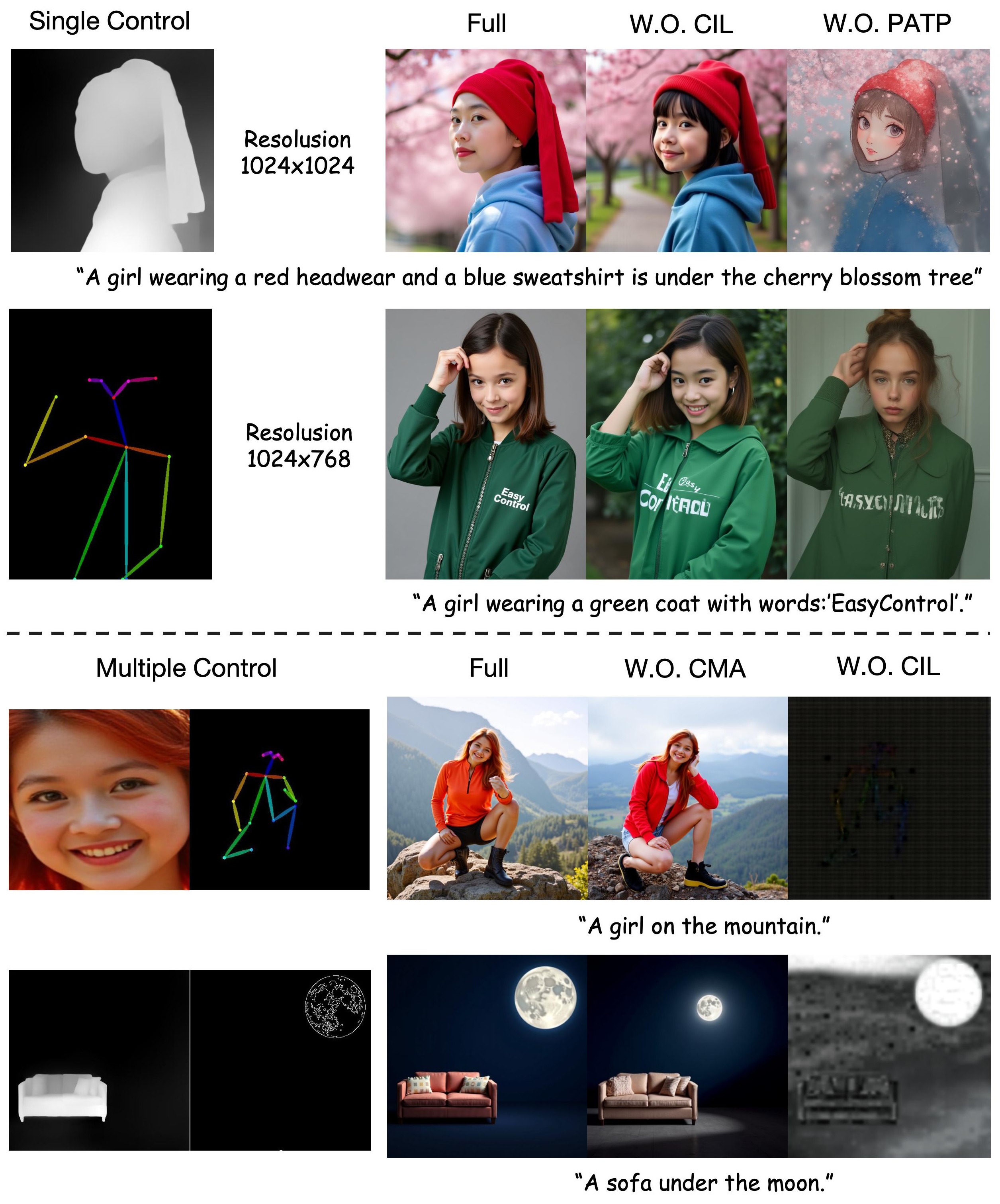}
    \vspace{-0.5cm}

    \caption{Visual ablation on different settings.}
    \vspace{-0.5cm}

    \label{fig:ab}
\end{figure}

Our model is trained exclusively on single-condition inputs, each condition token learns optimized interactions with denoising tokens.  During multi-condition inference, while all conditions interact normally with denoising tokens, inter-condition interference emerges due to untrained cross-condition token interactions (see Fig.~\ref{fig:ab}). This mechanism effectively integrates multiple conditional signals while avoiding interference during inference, Formally, we define the input sequence in multiple conditions inference as:
\begin{equation}
Z = [Z_{\text{t\&n}}; Z_{\text{c}_1}; Z_{\text{c}_2}; \dots; Z_{\text{c}_m}]
\end{equation}  
where \( Z_{\text{t\&n}}\) denotes the noise and text tokens, and \( Z_{\text{c}_i} \) represents the tokens corresponding to the \( i \)-th condition, we define an attention mask \( M \in \{0, -\infty\}^{n \times n} \) to regulate attention flow. Specifically, the mask is formulated as:  
\begin{equation}
M_{ij} =
\begin{cases}
0, & \text{if } i \in n_{\text{t\&n}} \text{ or } j \in n_{\text{t\&n}} \\
0, & \text{if } i, j \text{ belong to the same condition block} \\
-\infty, & \text{otherwise (cross-condition blocking)}
\end{cases}
\end{equation}
where \( n = n_{\text{t\&n}} + \sum_{i=1}^{m} n_{\text{c}_i} \) represents the total sequence length. This structured masking ensures that while the image tokens aggregate information from all conditions, distinct conditions remain isolated, preventing mutual interference.

\subsection{Position-Aware Training Paradigm}
\label{patp}
To improve computational efficiency and resolution flexibility in conditional image generation, we propose a Position-aware Training Paradigm. This paradigm is based on a naive approach: downscaling high-resolution control signals from their original dimensions $M \times N$ to a lower target resolution $H \times W$. In our experiments, we set $H = W = 512$. The resized image is then encoded into latent space via a VAE encoder, followed by a Patchify operation to extract Condition Tokens. These tokens are combined with Noise Tokens and Text Tokens from the original DiT model and processed through iterative denoising. 

While this naive downscaling approach works well for subject conditions (e.g., face images), it fails to preserve spatial alignment for spatial conditions (e.g., canny maps), limiting the model’s ability to generate images at arbitrary resolutions. To address this, we introduce two tailored strategies: (1) Position-Aware Interpolation (PAI) for spatial conditions, which maintains pixel-level alignment during resizing, and (2) PE Offset Strategy(see in supplementary Section B) for subject conditions, which applies a fixed displacement to position encodings in the height dimension.

\subsubsection{Position-Aware Interpolation}

To maintain spatial consistency between condition tokens and noise tokens, we introduce the Position-Aware Interpolation (PAI) strategy, which interpolates position encodings during the resizing process of conditional signals. This ensures the model accurately captures spatial relationships between control conditions and generated image pixels.

Given the original conditional image dimensions $(M, N)$ and the resized dimensions $(H, W)$, the scaling factors are computed as:
\begin{equation}
S_h = \frac{M}{H}, \quad S_w = \frac{N}{W}
\end{equation}
where $S_h$ and $S_w$ denote the scaling factors in height and width directions, respectively.

For a given patch $(i, j)$ in the resized conditional image, its corresponding position $(P_i, P_j)$ in the original image is mapped as:
\begin{equation}
P_i = i \times S_h , \quad P_j = j \times S_w
\end{equation}
where $i \in [0, H]$ and $j \in [0, W]$. This mapping aligns any patch in the resized image to its corresponding position in the original image.

The sequence of position encodings in the original image is represented as:
\begin{equation}
{(0,0), (0,1), \dots, (M,N)}
\end{equation}
while the interpolated sequence for the resized image is:
\begin{equation}
{(0,0), (0, S_w), \dots, (H \times S_h, W \times S_w)}
\end{equation}
This ensures that spatial relationships are preserved in the resized image.

\subsubsection{Loss Function}
Our loss function utilizes flow-matching loss. The mathematical expression is as follows:
\begin{equation}
L_{RF}=E_{t,\epsilon \sim N(0, I)}| | v_\theta(z, t, c_i)-\left(\epsilon-x_0\right)| |_2^2
\end{equation}
where \( z \) represents the image feature at time \( t \),$c_i$ is the input condition, \( v_\theta \) denotes the velocity field, \( x_0 \) refers to the original image feature, and \( \epsilon \) is the predicted noise.

\begin{figure*}[!t]
    \centering
         \vspace{-0.3cm}
    \includegraphics[width=1.\linewidth]{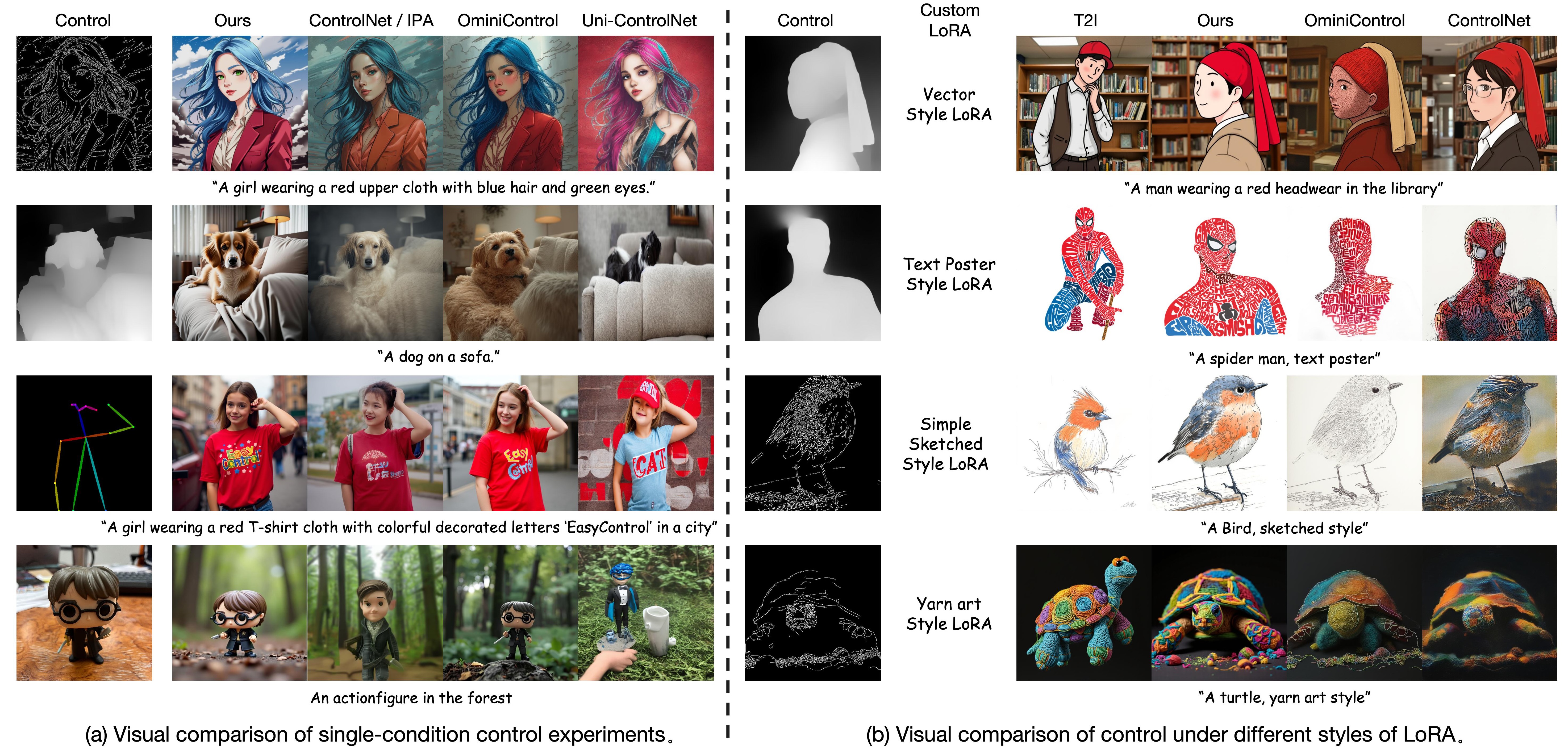}
     \vspace{-0.5cm}
    \caption{Visual comparison between different methods in single condition control. Figure (a) shows the results of each method under different control conditions and Figure (b) shows the adaptation of each method with different Style LoRA\cite{lora1,lora2,lora3,lora4} under control.}
        \vspace{-0.2cm}

    \label{fig:com1}
\end{figure*}

\begin{figure*}[!t]
    \centering
\vspace{-0.6cm}
\includegraphics[width=1.\linewidth]{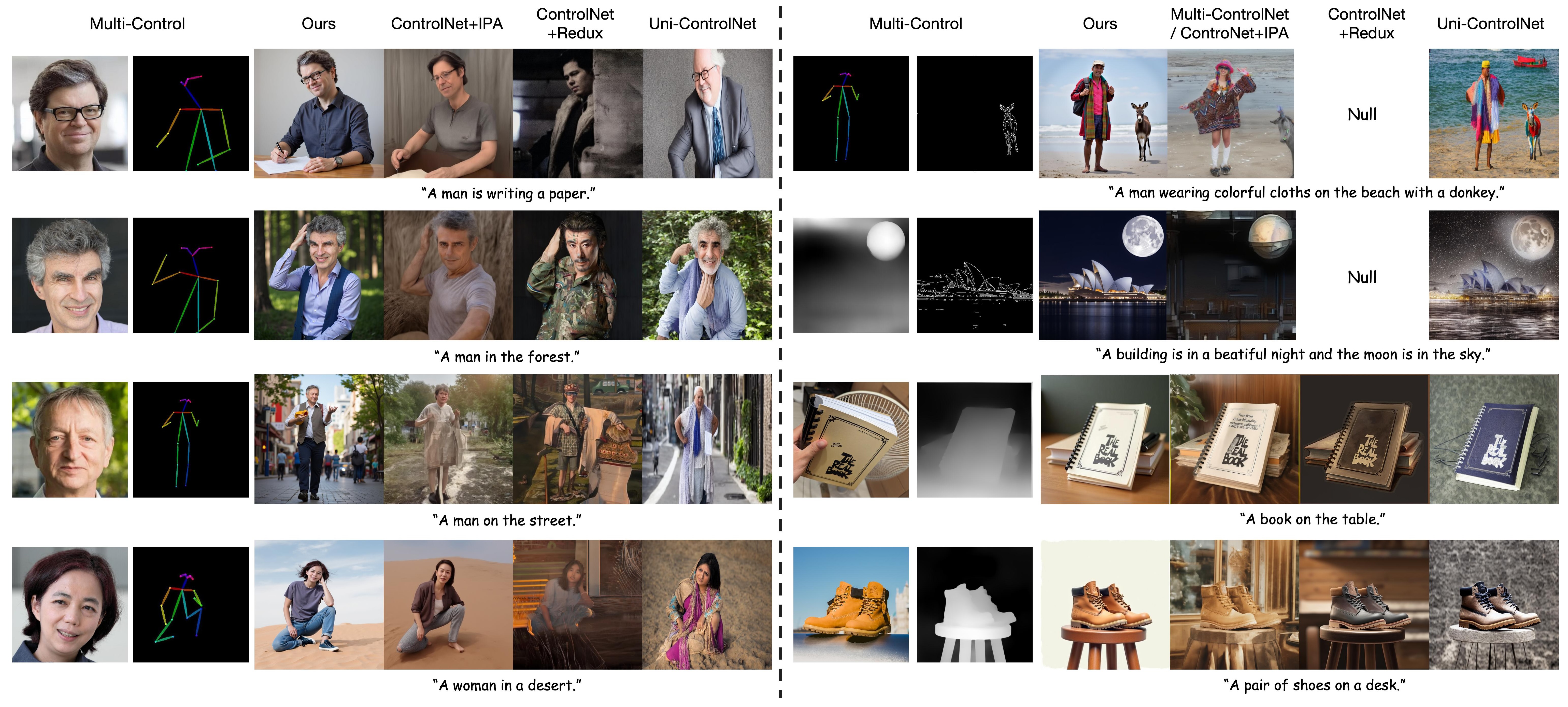}
    \vspace{-0.5cm}
    \caption{Visual comparison of different methods under multi-condition control.}
    \vspace{-0.3cm}
    \label{fig:com2}
\end{figure*}

\subsection{Efficient Inference via KV Cache}
\label{kvcache}
By leveraging the Causal Attention mechanism, our framework isolates the conditioning branch as a computation-independent module that remains agnostic to denoising timesteps. This unique design enables a novel application of the KV Cache technique during inference.

Since the conditioning branch’s computations are independent of the denoising timestep, we precompute and store the Key-Value (KV) pairs of all conditional features only once during the initial timestep. These cached KV pairs are reused across all subsequent timesteps, eliminating redundant recomputation of identical conditional features. This approach reduces inference latency by avoiding $N$-fold recomputation (for $N$ denoising steps) while preserving generation quality and model flexibility (detailed in supplementary Sec.~D).


\section{Experiments}
\newcommand{\best}[1]{\textbf{#1}}

This section begins with a description of the implementation details of EasyControl, followed by an overview of the evaluation metrics. We then present the experimental results, including both qualitative and quantitative analyses, as well as ablation experiments.

\subsection{Implementation Details} 
We employed FLUX.1 dev as the pre-trained DiT. For each spatial or subject condition training, we utilize 4 A100 GPUs(80GB), a batch size of 1 per GPU, and a learning rate of 1e-4, training over 100,000 steps. During inference, the flow-matching sampling is applied with 25 sampling steps. (Training Data is detailed in supplementary Sec.~C.)
\setlength{\tabcolsep}{1.5mm}{

\begin{table}[!t]
  \small
  \centering
  \begin{tabular}{c|c|c|c}
\toprule
\multirow{2}{*}{Cond.} & \multirow{2}{*}{Method} &\multirow{2}{*}{Time(s) $\downarrow$} &\multirow{2}{*}{Params$\downarrow$} \\ 
& &  & \\
\midrule
\midrule
& ControlNet &\underline{16.5}& 3B \\
& OminiControl &31.6 &15M \\
\cdashline{2-4}
\footnotesize{Single}&Ours(w.o. PATP\&KVCache) &38.9 &- \\
&Ours(w.o. KVCache) & 22.4\textcolor{red}{(-42\%)} &- \\
& Ours(w.o. PATP) &25.0\textcolor{red}{(-36\%)} &- \\
& \cellcolor{cvprblue!15}Ours(Full) &\cellcolor{cvprblue!15}\best{16.3}\textcolor{red}{(-58\%)} &\cellcolor{cvprblue!15}\best{15M} \\

\midrule
 & ControlNet+IPA &\best{16.8} &4B \\
& Multi-ControlNet &20.3 &6B \\
\cdashline{2-4}
\footnotesize{Double}&Ours(w.o. PATP\&KVCache) &72.4 &- \\
& Ours(w.o. KVCache) &29.9\textcolor{red}{(-59\%)} &-  \\
& Ours(w.o. PATP) &36.7\textcolor{red}{(-50\%)} &- \\
& \cellcolor{cvprblue!15}Ours(Full) &\cellcolor{cvprblue!15}\underline{18.3\textcolor{red}{(-75\%)}} &\cellcolor{cvprblue!15}\best{30M} \\

\bottomrule
\end{tabular}
\caption{Quantitative efficiency comparison with baseline methods and different settings. The inference time is calculated when generating a 1024$\times$1024 resolution image with 25 denoising steps. The parameters refer exclusively to those of the additional module, excluding the parameters of the base model.}
\vspace{-0.5cm}
\label{tab:effeciency}
\end{table}
}

\subsection{Experimental Settings}
\noindent\textbf{Visual Comparisons}:
We evaluate the following settings:
(1) Single-condition generation,
(2) Single-condition adaptation with customized models,
(3) Multi-condition integration (illustrated in Figures~\ref{fig:com1} and~\ref{fig:com2}, we also compare with several ID customization methods \cite{instantid,li2024photomaker,he2024uniportrait} in supplementary Sec.~F for details), and
(4) Resolution adaptability (detailed in supplementary Sec.~G).
\textbf{Quantitative Comparisons}:
We assess the following aspects:
(1) Inference time and model parameter count under single- and dual-condition generation (to evaluate efficiency, as shown in Table~\ref{tab:effeciency}),
(2) Controllability, generation quality, and text consistency using face + OpenPose as multi-conditions (detailed in supplementary Sec.~F), and
(3) Controllability, generation quality, and text consistency under single-condition settings (detailed in supplementary Sec.~E).

\noindent\textbf{Comparison Methods}
For single-condition, we compare with Controlnet\cite{controlnet}, OminiControl\cite{tan2024ominicontrol}, and Uni-ControlNet\cite{zhao2024uni}. For multi-condition settings, we evaluate our approach against several plug-and-play baseline methods including \textit{Controlnet+IP-Adapter}\cite{ipadapter}, \textit{Controlnet+Redux\cite{redux}}, and \textit{Uni-Controlnet}\cite{zhao2024uni}. We also compare several ID customization methods\cite{li2024photomaker,instantid,he2024uniportrait} integrated with ControlNet.
\subsection{Experiment Results}
\subsubsection{Qualitative Comparison}
Fig.~\ref{fig:com1} (a) compares the performance of different methods under single-control conditions. Under Canny control, Uni-ControlNet and ControlNet exhibit color inconsistencies, causing deviations from the input text. Under Depth Control, Uni-ControlNet fails to generate coherent images, while ControlNet and OmniControl introduce artifacts, such as the fusion of the dog and the sofa. Under OpenPose control, our method preserves text rendering, whereas others weaken or lose this ability. In Subject Control, IP-Adapter and Uni-ControlNet fail to align with the reference. Overall, our method ensures text consistency and high-quality generation across diverse control conditions.

Fig~\ref{fig:com1} (b) compares the Plug-and-play capability of different methods in generating images on four custom models. The leftmost column shows the original text-to-image (T2I) results from the LoRA fine-tuned Flux.1 Dev model. Both ControlNet and OminiControl sacrifice stylization, and suffer from quality degradation. In contrast, our method demonstrates the ability to minimize stylization loss without losing controllability which demonstrates the plug-and-play capability of our method.

Fig.~\ref{fig:com2} presents a visual comparison of different methods under multi-condition control. For OpenPose and Face, our approach achieves superior identity consistency and controllability. In contrast, other methods exhibit conflicts between control conditions. While the combination of ControlNet and IP-Adapter maintains controllability, it compromises identity consistency. ControlNet+Redux and Uni-ControlNet fail to preserve both identity consistency and controllability, which is also observed in subject-depth control scenarios (Right third/fourth row). For OpenPose-Canny and Depth-Canny combinations, both our method and Uni-ControlNet generate images that satisfy the control conditions. However, Uni-ControlNet struggles to align with textual inputs and produces lower-quality images. Multi-ControlNet fails to satisfy both conditions simultaneously. These results demonstrate the flexibility of our method in seamlessly integrating multiple conditions.

\subsubsection{Quantitative Comparison}
Table~\ref{tab:effeciency} presents the inference time and corresponding model parameter counts for various algorithms on a single A100 GPU with 20 sampling steps. In single-condition settings, our full model achieves the best performance with an inference time of 16.3 seconds, representing a \best{58\%} reduction compared to the ablated version without the Position-Aware Training Paradigm (PATP) and KV Cache. Notably, our method achieves this efficiency while maintaining a minimal parameter count of 15M, which is significantly lower than ControlNet's 3B parameters. For double-condition tasks, our full model achieves an inference time of 18.3 seconds, which is \best{75\%} faster than the ablated version without PATP and KV Cache. This performance is competitive with ControlNet+IPA (16.8 seconds) while maintaining a much smaller model size (30M parameters compared to 4B for ControlNet+IPA). The results highlight the effectiveness of our proposed PATP and KV Cache mechanisms in improving inference efficiency without compromising model compactness. 

\subsubsection{Ablation Study}
In our ablation study, we analyze the impact of removing each module. First, replacing the Condition Injection LoRA (CIL) with the standard LoRA structure (W.O. CIL) enables single-condition control but fails to generalize to multi-condition control in a zero-shot manner. For the Position-Aware Training Paradigm (PATP), we trained a model W.O. PATP, where both the control signal and noise were fixed at 512×512 resolution while keeping other training settings unchanged. This model exhibits artifacts and quality degradation when generating high-resolution (e.g., 1024×1024) or non-square aspect ratio (e.g., 1024×768) images. In contrast, our PATP-based training effectively mitigates these issues. For Causal Attention, removing Causal Mutual Attention(CMA) still allows image generation due to the adaptive nature of attention. However, conflicts between conditions reduce control accuracy, leading to deviations such as altered human poses in Multi-Control scenarios and shifts in object positions, such as the moon. When all modules are used together, our method achieves the highest controllability, generation quality, and adaptability to varying resolutions and aspect ratios.
\section{Conclusion}
In conclusion, we present EasyControl, a highly efficient and flexible framework for unified condition-guided diffusion models. Our framework leverages three key innovations:(1) A lightweight Condition Injection LoRA Module, which enables seamless integration of diverse condition signals without altering the core model's functionality. (2) A Position-Aware Training Paradigm, which ensures adaptability to various resolutions and aspect ratios. (3) A novel Causal Attention Mechanism combined with the KV Cache technique, which significantly improves efficiency. Together, these components address the challenges of efficiency and flexibility in controllable image generation. EasyControl achieves strong controllability and high-quality results across a wide range of visual tasks. Extensive experiments demonstrate its ability to handle complex, multi-condition scenarios while scaling to diverse resolutions and aspect ratios. Our framework offers a powerful and adaptable solution for conditional image generation.
\clearpage
\noindent\textbf{\Large Appendix.}
\renewcommand\thesection{\Alph{section}}
\setcounter{section}{0}

\section{Preliminary: Diffusion Transformer}
\label{Preliminary}
Currently, The state-of-the-art text-to-image Diffusion model architectures are based on Diffusion Transformers (DiT)\cite{dit}, including models such as SD3\cite{sd3}, FLUX\cite{flux2024}. These models integrate diffusion processes with Transformer architectures to improve text-to-image generation, yielding high-quality and accurate text-to-image synthesis.

Our approach is based on the FLUX.1 pre-trained model, which consists of three key components: T5 as the text encoder, a VAE for image compression, and a Transformer-based denoising network. Specifically, The denoising network divides the latent noise into several patches and treats each patch as a noise token, denoted as \( X \in \mathbb{R}^{N \times d} \), where \( N \) is the number of noise tokens and \( d \) is the dimensionality of each token. To effectively introduce spatial position information in different noise patches, FLUX.1 employs rotary position encoding (RoPE)\cite{rope} to encode spatial information within each noise patch. Meanwhile, the text prompts are encoded into text tokens \( C_T \in \mathbb{R}^{M \times d} \) through the T5 text encoder, where \( M \) represents the number of text tokens. These image and text tokens are then fused by concatenation to form a joint representation. After feature fusion, the image-text token sequence is fed parallel into a transformer-based denoising model. The model iteratively denoises the image, progressively restoring a clear image and ultimately generating a high-quality image that aligns with the textual description.

\section{Position Encoding Offset}
\label{peo}
The PE offset strategy was proposed by method\cite{tan2024ominicontrol}, which applies a fixed displacement to position encodings and was proved to lead to faster convergence. This offset is uniform across all encodings within the subject condition image. In our experiments, we set this offset to 64 in the height dimension. Mathematically, for each position encoding $PE(i, j)$ in the subject condition image, the adjustment is:
\begin{equation}
PE(i, j) \leftarrow PE(i, j) + \Delta_h \cdot \mathbf{e}_h
\end{equation}
where $\mathbf{e}_h$ is the unit vector along the height dimension, and $\Delta_h = 64$ ensures distinct separation between spatial and subject conditions.

\section{Trianing Data}
\label{trainingdata}
\begin{figure}[!t]
    \centering
    \includegraphics[width=1.\linewidth]{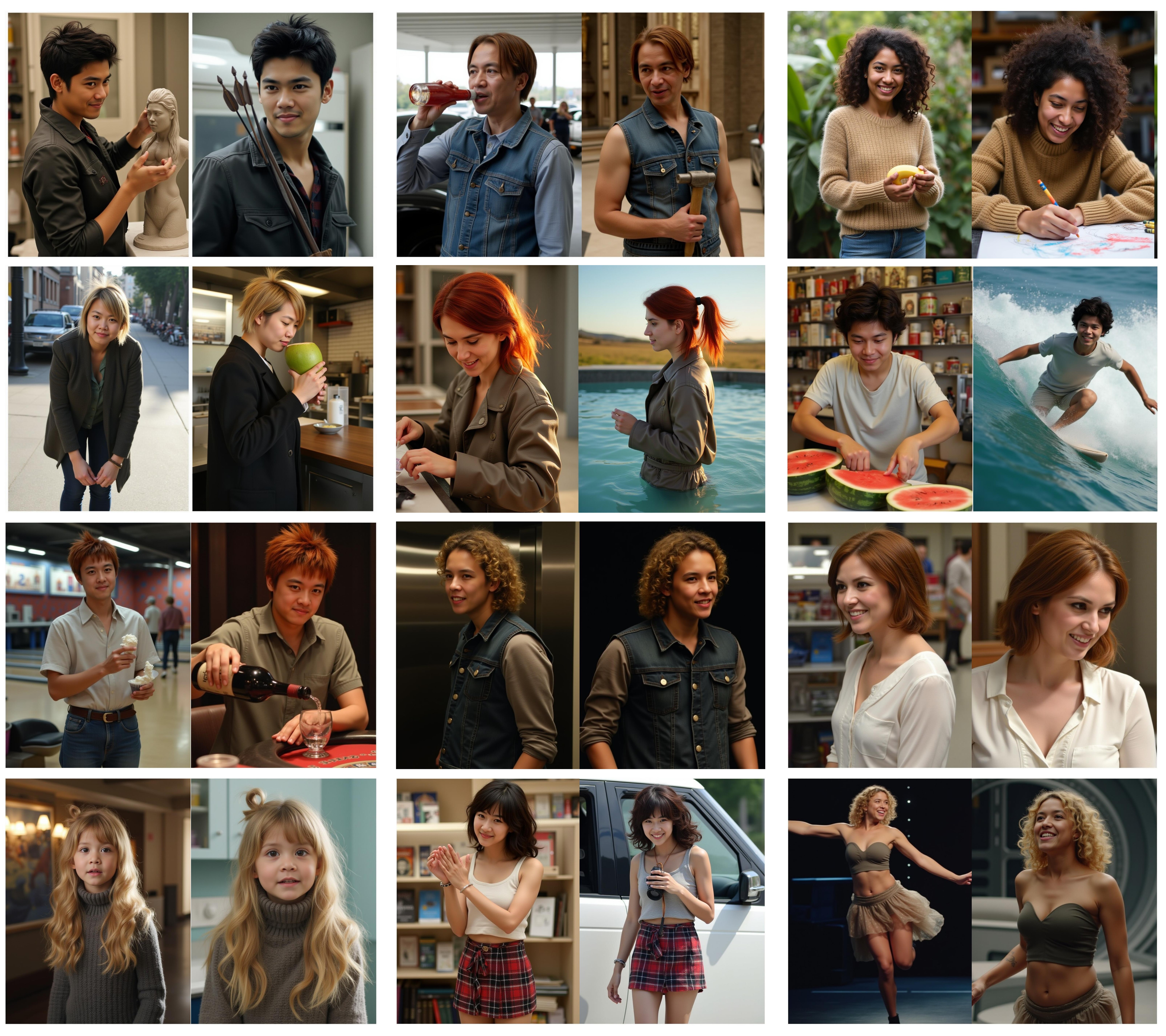}
    \caption{Visualization of samples in private Multi-view Human Dataset.}
    \label{fig:data}
\end{figure}

For spatial control tasks such as depth, canny, and OpenPose, we employ the MultiGen-20M dataset\cite{zhao2024uni} as our primary training resource. Regarding subject control, our training is conducted using the Subject200K dataset\cite{tan2024ominicontrol}. For face control, we utilize a curated subset of the LAION-Face dataset\cite{laionface}, supplemented by a collected private multi-view human dataset (See Fig~\ref{fig:data}), where all human images are preprocessed through InsightFace\cite{deng2018arcface} for precise cropping and alignment to ensure consistency and accuracy in our training inputs.

\begin{algorithm}[t]
\caption{KV Cache for Efficient Conditional Image Generation}
\label{alg:kvcache}
\begin{algorithmic}[1]
\REQUIRE 
    \STATE Conditional features $\{\text{cond}_i\}_{i=1}^m$ for $m$ conditions
    \STATE Denoising steps $T = \{t_0, t_1, ..., t_T\}$
\ENSURE Generated image $x_0$ with reduced latency
\STATE Initialize KV cache dictionary $\mathcal{D} \gets \emptyset$
\STATE Generate initial noisy image $x_N \sim \mathcal{N}(0,I)$

\FOR{timestep $t \in \{t_T, t_{T-1}, ..., t_0\}$}
    \IF{$t = t_T$ (First step)}
        \FOR{each condition branch $i \in \{1,...,m\}$}
            \STATE Compute keys/values: $K_{C_i}, V_{C_i} = f_{\theta}(\text{cond}_i)$
            \STATE Cache KV pairs: $\mathcal{D}[i] \gets (K_{C_i}, V_{C_i})$
        \ENDFOR
    \ENDIF
    
    \STATE Retrieve cached KV pairs: $\{(K_{C_i}, V_{C_i})\}_{i=1}^m \gets \mathcal{D}$
    \STATE Compute self-attention using current noise and text features: $Q_{\text{denoising}}, K_{\text{denoising}}, V_{\text{denoising}} = f_{\theta}(\text{x}_t, t)$
    \STATE Fuse conditions via cached K/V pairs:
    \[
    Q = Q_{\text{denoising}}, \]
    \[K = \text{Concat}\left(K_{\text{denoising}}, K_{C_1}, \dots, K_{C_m}\right), \]
    \[V = \text{Concat}\left(V_{\text{denoising}}, V_{C_1}, \dots, V_{C_m}\right)
    \]
    \[
    \text{Output} = \text{Softmax}\left(\frac{QK^\top}{\sqrt{d}}\right)V
    \]
    \STATE Update latent: $x_{t-1} = \text{Denoise}(x_t, t, \text{Output})$
\ENDFOR

\STATE \textbf{return} Final image $x_0$
\end{algorithmic}
\end{algorithm}

\section{Details of KV Cache} 
\label{sup:kvcache}
More details of KV Cache for Efficient Conditional Image
Generation is shown in the algorithm~\ref{alg:kvcache}.

\section{Single Condition Quantitative Comparison}
\label{singlecom}

\setlength{\tabcolsep}{4.5mm}{
\begin{table*}[!h]
  \small
  \centering
  \begin{tabular}{c|l|c|cc|c}
\toprule
\multirow{2}{*}{Condition} & \multirow{2}{*}{Method} & Controllability & \multicolumn{2}{c|}{Generative Quality} & Text Consistency \\
&   & $\text{F1} \uparrow / \text{MSE} \downarrow$ & $\text{FID} \downarrow$ & $\text{MAN-IQA} \uparrow$ & $\text{CLIP-Score} \uparrow$ \\ 
\midrule
\midrule
\multirow{4}{*}{Canny} & ControlNet &0.232&20.325&0.420 &0.271\\
&  OminiControl &\best{0.314} &\underline{17.237}&\underline{0.471}&\underline{0.283}\\
& Uni-ControlNet &0.201&17.375&0.402&0.279\\
& Ours\cellcolor{cvprblue!15} & \cellcolor{cvprblue!15} \underline{0.311}&\cellcolor{cvprblue!15}\best{16.074} &\cellcolor{cvprblue!15}\best{ 0.503} &\cellcolor{cvprblue!15} \best{0.286}\\

\midrule

\multirow{4}{*}{Depth} & ControlNet &1781&23.968&0.319 &0.265\\
& OminiControl & \underline{1103} &\best{18.536}&\underline{0.431}&\underline{0.285}\\
& Uni-ControlNet &1685&21.788&0.423&0.279\\
& \cellcolor{cvprblue!15}  Ours &\cellcolor{cvprblue!15}\best{1092} & \cellcolor{cvprblue!15} \underline{20.394}& \cellcolor{cvprblue!15} \best{0.469} & \cellcolor{cvprblue!15} \best{0.289}\\

\bottomrule
\end{tabular}
\caption{Quantitative comparison with baseline methods on single condition tasks.}
\label{tab:single_results1}
\end{table*}
}

\setlength{\tabcolsep}{4mm}{
\begin{table*}[!h]
  \small
  \centering
  \begin{tabular}{c|c|cc|cc|c}
\toprule
\multirow{2}{*}{Condition} & \multirow{2}{*}{Method} & \multicolumn{2}{c}{Identity Preservation} & \multicolumn{2}{c}{Generative Quality} & Text Consistency \\
& & $\text{CLIP-I} \uparrow$ & $\text{DINO-I} \uparrow$ & $\text{FID} \downarrow$ & $\text{MAN-IQA} \uparrow$ & $\text{CLIP-Score} \uparrow$ \\ 

\midrule
\multirow{4}{*}{Subject} & IP-Adapter 
& \textbf{0.700} & 0.429 & 79.277 & 0.511 & 0.266 \\
& OminiControl & 0.663 & \textbf{0.445} & \underline{72.298} & \underline{0.579} & \underline{0.276} \\
& Uni-ControNet & 0.641 & 0.417& 86.369 & 0.439 & 0.204\\
& \cellcolor{cvprblue!15}Ours  & \cellcolor{cvprblue!15}\underline{0.667} &\cellcolor{cvprblue!15}\underline{0.443} &\cellcolor{cvprblue!15}\textbf{71.910} &\cellcolor{cvprblue!15}\textbf{0.595} & \cellcolor{cvprblue!15}\textbf{0.283} \\
\bottomrule
\end{tabular}
\caption{Quantitative comparison with baseline methods on single condition tasks.}
\label{tab:single_results2}
\end{table*}

\setlength{\tabcolsep}{2mm}{
\begin{table*}[!h]
  \small
  \centering
  \begin{tabular}{c|c|c|c|cc|c}
\toprule
\multirow{2}{*}{Condition} & \multirow{2}{*}{Method} &ID Preservation &Controllability & \multicolumn{2}{c|}{Generative Quality} & Text Consistency\\
& &$\text{Face Sim.} \uparrow$ & $\text{MJPE}\downarrow$ & $\text{FID} \downarrow$ & $\text{MAN-IQA} \uparrow$ & $\text{CLIP-Score} \uparrow$ \\ 
\midrule
\midrule

\multirow{7}{*}{Openpose+Face} & ControlNet+IPA & 0.049 &166.7 &227.06 &0.229 &0.156\\
& ControlNet+Redux & 0.027 & 141.5&\underline{200.70} &0.293 &0.217\\
& Uni-ControlNet &0.048 &258.8 &203.31 &0.481 &0.147\\

& ControlNet+InstantID & \underline{0.521} &83.9 & 203.17 &0.345& 0.250\\
& ControlNet+PhotoMaker & 0.343 & 86.3 &213.83 &0.420  &\underline{0.281}\\
& ControlNet+Uni-portrait & 0.456 & \underline{46.0} &203.07 & \underline{0.564} &0.253\\
& \cellcolor{cvprblue!15}Ours & \cellcolor{cvprblue!15}\textbf{0.530} & \cellcolor{cvprblue!15}\best{36.7} &\cellcolor{cvprblue!15}\best{184.93} &\cellcolor{cvprblue!15}\best{0.586} &\cellcolor{cvprblue!15}\best{0.285}\\

\bottomrule
\end{tabular}
\caption{Quantitative comparison with baseline methods on multi-condition tasks.}
\label{tab:idresults}
\end{table*}
}

\textbf{Settings.}
In this section, we compare our method with \textit{Controlnet}\cite{controlnet}, \textit{OmniControl}\cite{tan2024ominicontrol}, and \textit{Uni-Control}\cite{zhao2024uni} using two types of conditioning: Depth and Canny. For the subject condition, We compare our method with \textit{OmniControl}\cite{tan2024ominicontrol}, \textit{IP-Adapter(IPA)}\cite{ipadapter}, and \textit{Uni-ControlNet}\cite{zhao2024uni}. (To ensure a fair and consistent comparison, all methods are implemented using FLUX.1 dev as the base model, with configurations and parameters sourced from publicly available official\cite{controlnet,redux,tan2024ominicontrol} and community resources\cite{ControlNet_union,IPA_xlab} with recommended parameters, while Uni-ControlNet employs its official implementation\cite{zhao2024uni} based on the SD1-5 architecture.)

\textbf{Data.}
For the Depth map and Canny edge conditions, comprehensive evaluations were conducted on the COCO 2017\cite{coco} validation set comprising 5,000 images. All generated outputs strictly preserved the original image resolutions and aspect ratios, with textual prompts derived from the corresponding ground-truth captions of the dataset. For subject control scenarios, we adopted the Concept-101 benchmark dataset\cite{customdiff} to assess model performance. Each reference image was paired with its semantically aligned textual description as the conditioning prompt.

\textbf{Metrics.} To comprehensively evaluate the performance of each algorithm, we assess four key aspects: \textit{1. Controllability:} We extract structural information from the generated images using the corresponding structure extractor, obtaining the structure map. The F1 Score is computed between the extracted and input edge maps in the edge-conditioned generation, and the MSE is calculated between the extracted and original condition maps for the depth task. \textit{2. Text Consistency:} We use the CLIP-Score\cite{CLIP,hessel2021clipscore} to evaluate the consistency between the generated images and the input text. \textit{3. Generative Quality:} The diversity and quality of the generated images are assessed using FID\cite{fid}, MAN-IQA\cite{maniqa}.\textit{4. Identity Preservation:} For the subject condition, we use CLIP-I\cite{CLIP} and DINO-I\cite{dino} to evaluate identity preservation. Specifically, CLIP-I computes the cosine similarity between image embeddings extracted by the CLIP image encoder for both generated and reference images. Similarly, DINO-I measures identity preservation by calculating the cosine similarity of image embeddings obtained through the DINO encoder framework. 

\textbf{Quantitative Analysis.} As shown in Table~\ref{tab:single_results1}, our proposed method demonstrates superior performance across multiple evaluation metrics. Under the Canny condition, our approach outperforms all comparative methods in terms of generation quality and text consistency, while achieving the second-highest score in controllability. In the depth condition, our method exhibits dominant performance in both controllability and text consistency. Regarding generation quality, while our method ranks second position in the FID metric, it achieves first according to the MAN-IQA metric. These comprehensive results substantiate the superiority of our approach across most evaluation criteria, particularly highlighting its exceptional performance in controllability, generation quality, and text consistency. As shown in Table~\ref{tab:single_results2}, Under the Subject condition, our approach outperforms all comparative methods in terms of generation quality and text consistency and achieves competitive results on identity preservation.

\textbf{Qualitative Analysis.}
We show some results about spatial control in figure~\ref{fig:sup_spatial}. Under identical conditional input configurations, both ControlNet and OminiControl demonstrate significant blurring artifacts in the synthesized images. In contrast, our framework consistently preserves superior visual fidelity across all evaluated scenarios. This qualitative advantage is particularly pronounced in the preservation of fine-grained details and structural integrity, thereby substantiating the enhanced performance of our Position-Aware Training Paradigm. We have also visualized several subject control results in Figure~\ref{fig:sup_subject} to demonstrate the effectiveness of our method in terms of identity preservation, generative quality, and text consistency. (The prompts utilized in the generated images include: \textit{in the forest, in the library, on a snow-covered mountain, in the city, in a room, in front of a castle, floating on water, on the beach, on a mountain, in the desert, and on a snowy day}.)

\section{Multi-Condition Quantitative Comparison}
\label{multicom}

\textbf{Settings.}
In this section, we conduct comparisons using face + OpenPose as multi-condition configurations, against several plug-and-play baseline methods including: \textit{Controlnet+IP-Adapter}\cite{ipadapter}, \textit{Controlnet+Redux\cite{redux}}, and \textit{Uni-Controlnet}\cite{zhao2024uni} and several SOTA play-and-plug identity customization methods, including \textit{Controlnet+InstantID\cite{instantid}}, \textit{Controlnet+PhotoMaker\cite{li2024photomaker}}, and \textit{ControlNet+Uni-portrait\cite{he2024uniportrait}}. (For several plug-and-play baseline methods, we utilize official FLUX-based implementations such as OminiControl and community-driven implementations including ControlNet, IPA, and Redux. For identity customization methods, we adopt official implementations based on SD1-5 (e.g., Uni-portrait) and SDXL (e.g., PhotoMaker, InstantID) as base models, along with their corresponding ControlNet modules.)

\textbf{Data.}
For evaluation, we constructed a comprehensive dataset comprising three components: (1) 1,000 randomly sampled face images from the FFHQ dataset for face control inputs; (2) 1,000 full-body or half-body human images crawled from Laion face dataset\cite{laionface}, from which OpenPose information was extracted for spatial control inputs; and (3) 1,000 text prompts generated by GPT, each describing a person with specific characteristic and locations. Each algorithm generated 1,000 images based on these inputs for evaluation. This diverse dataset ensures a thorough assessment of the models' capabilities in handling various control conditions.

\textbf{Metrics.}
To comprehensively evaluate the performance of each algorithm, we assess several key aspects: \textit{1. Controllability:} The Mean Joint Position Error (MJPE) metrics computed between the extracted and input openpose maps in the pose generation. \textit{2. Text Consistency:} We use the CLIP-Score to evaluate the consistency between the generated images and the input text. \textit{3. Generative Quality:} The diversity and quality of the generated images are assessed using FID\cite{fid}, MAN-IQA\cite{maniqa}. \textit{4. Identity Preservation:} For the face condition, we use face similarity\cite{deng2018arcface} to evaluate identity preservation. For the OpenPose condition, our controllability metric is quantitatively assessed through the Mean Joint Position Error (MJPE) metrics, which measure the spatial consistency between the generated image and the input OpenPose map. The evaluation procedure involves three sequential steps: initially, key point information is extracted from both the generated image and the input condition using OpenPose. Subsequently, The Euclidean distance for each joint is computed and averaged to obtain the MJPE for a single image. This process is repeated for all images in the test set, and the average MJPE serves as the model's controllability metric. By quantifying joint position deviations, MJPE effectively evaluates the consistency between generated images and input conditions. It is noteworthy that a lower MJPE indicates superior spatial alignment and better pose consistency between the generated image and the input condition, thus reflecting higher controllability in the pose generation process.

\subsection{Quantitative Comparison}
The quantitative results are presented in Table~\ref{tab:idresults}. Our method achieves state-of-the-art performance across all metrics. Specifically, it obtains the best Face Similarity, demonstrating superior ID preservation. For controllability, our approach achieve the lowest MJPE score and significantly outperforms others. In terms of generative quality, our method achieves the lowest FID and highest MAN-IQA, indicating better image quality and diversity. Additionally, it maintains strong text consistency with the highest CLIP score. These results collectively demonstrate the effectiveness of our framework in balancing control precision, identity preservation, and generation quality under a multi-condition combination.  

It is noteworthy that in Table~\ref{tab:idresults}, certain algorithms exhibit significantly inferior performance in terms of Face Similarity (Face Sim) and Mean Joint Position Error (MJPE) metrics compared to other methods. This is primarily attributed to the fact that many competing methods fail to effectively transfer facial or pose features from the control images, often resulting in generated images that are blurry, distorted, or lack detectable facial or pose features. Consequently, these methods are unable to accurately compute the metrics required for face similarity or pose alignment. In contrast, our approach ensures robust feature transfer and precise alignment, enabling the generation of high-quality images with clearly detectable facial and pose attributes, which contributes to the superior performance reflected in the metrics.

\subsection{Visual Comparison}
As illustrated in the figure~\ref{fig:sup_fo}, we present a visual comparison with ID customization methods. Our method demonstrates superior performance in facial similarity, controllability, and image quality compared to other approaches. This indicates that our framework, despite being trained on single conditions, exhibits strong plug-and-play adaptability, effectively integrating multiple conditions without conflicts. In contrast, other methods often suffer from incompatibility between different modules, leading to degraded facial similarity, controllability, and poor generation quality. The visual results further validate the robustness and versatility of our approach in handling complex multi-condition generation tasks.

\section{Visual Comparison of Resolution Adaptability}
\label{res}
As shown in the figure~\ref{fig:sup_res}, we compare the controllability of our method with DiT-based controllable baseline methods, including ControlNet and OmniControl, across different resolutions. Clearly, our approach consistently demonstrates strong controllability, high text consistency, and superior image quality across resolutions ranging from low to high. However, at lower resolutions, ControlNet exhibits image distortion, while at higher resolutions, OmniControl also suffers from image degradation. This demonstrates that our method exhibits strong adaptability across different resolutions.

\section{Limitations}
While the proposed framework demonstrates significant improvements in flexibility and computational efficiency compared to existing DiT-based approaches, certain technical limitations remain and warrant further investigation. For example, in multi-conditional scenarios involving conflicting inputs, the model may generate artifacts characterized by overlapping layers, as illustrated in Figure~\ref{fig:sup_lim}. Additionally, our method cannot indefinitely upscale generated resolutions. When the resolution becomes extremely high, there is a decrease in the ability to control the output.

\begin{figure}[!t]
    \centering
\includegraphics[width=1.\linewidth]{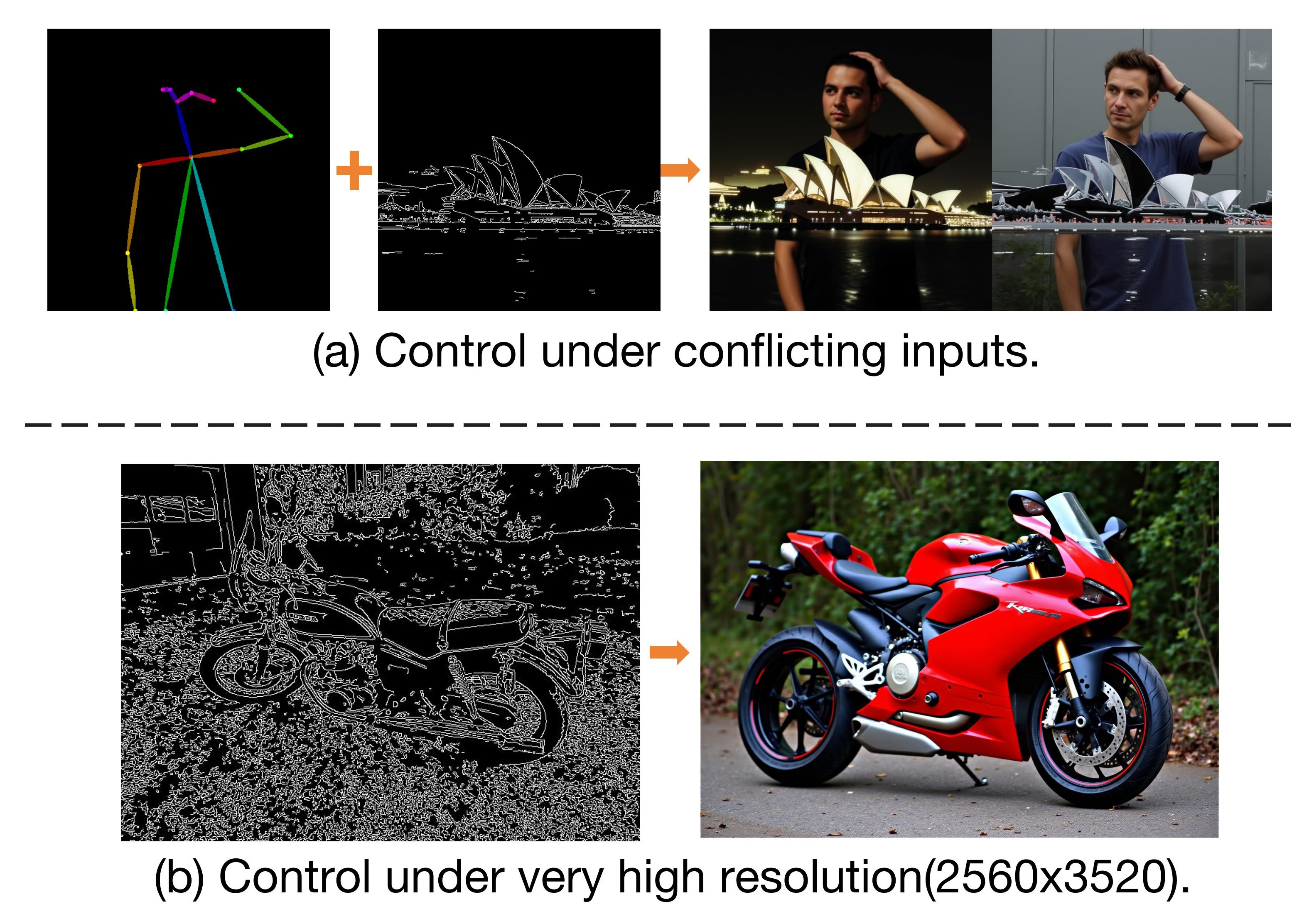}
    \caption{Visualization of results (1) under conflicting condition inputs (2) under very high-resolution generation.}
    \label{fig:sup_lim}
\end{figure}

\begin{figure*}[!b]
    \centering
    \includegraphics[width=1.\linewidth]{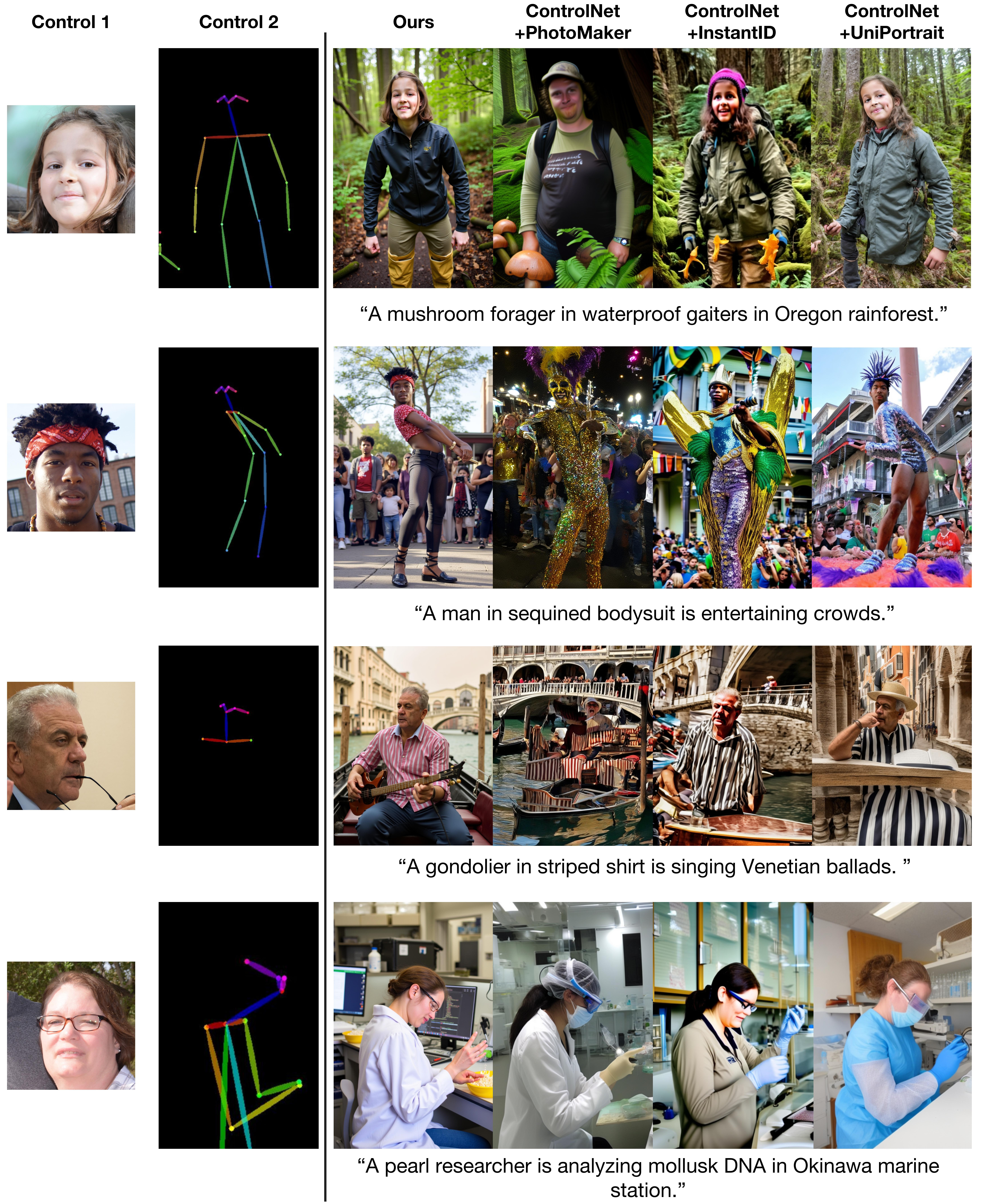}
    \caption{Visual comparison with Identity customization methods under multi-condition generation setting.}
    \label{fig:sup_fo}
\end{figure*}

\begin{figure*}[!b]
    \centering
    \includegraphics[width=1.\linewidth]{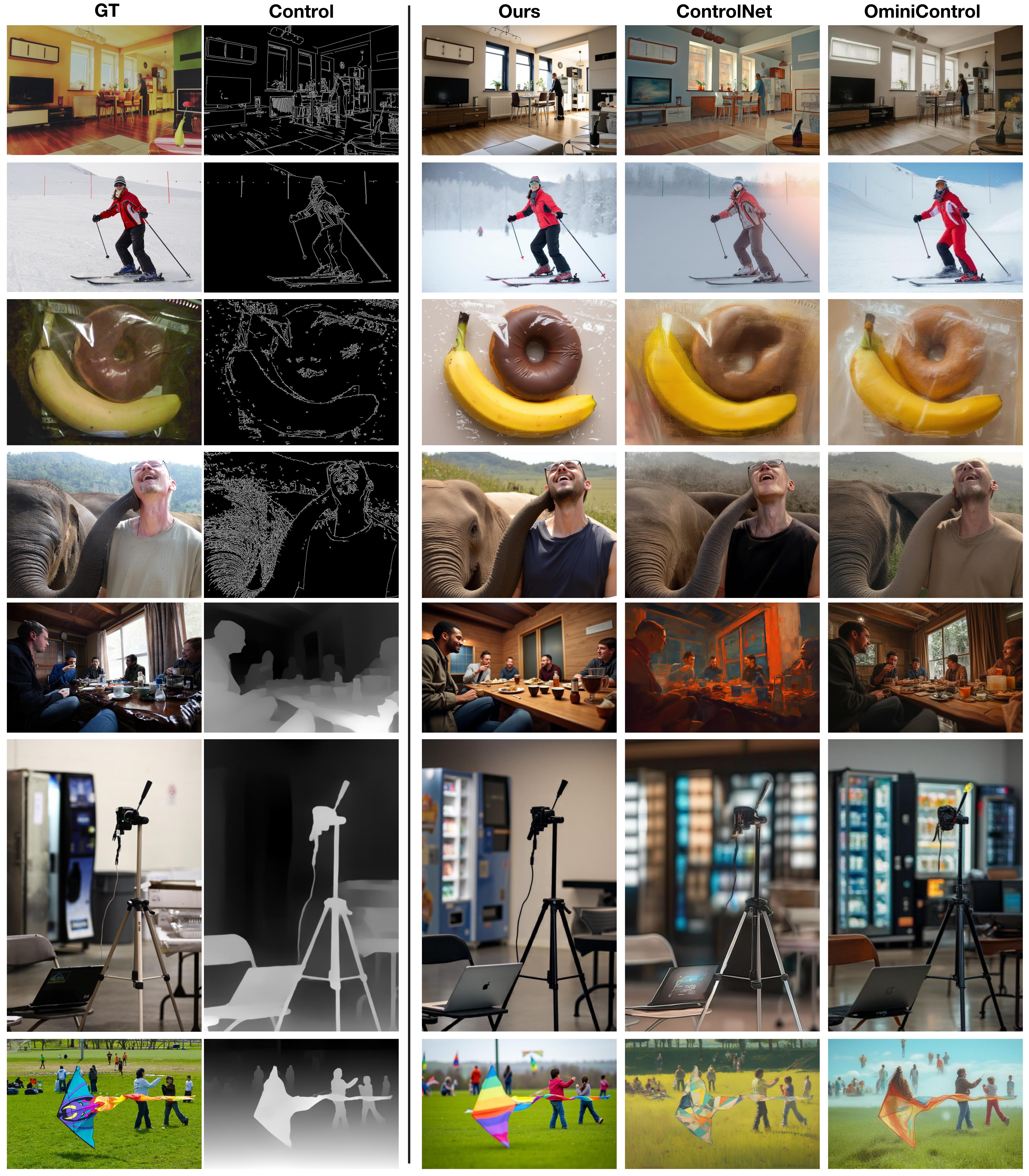}
    \caption{Visualization of spatial control generation.}
    \label{fig:sup_spatial}
\end{figure*}

\begin{figure*}[!b]
    \centering
    \includegraphics[width=1.\linewidth]{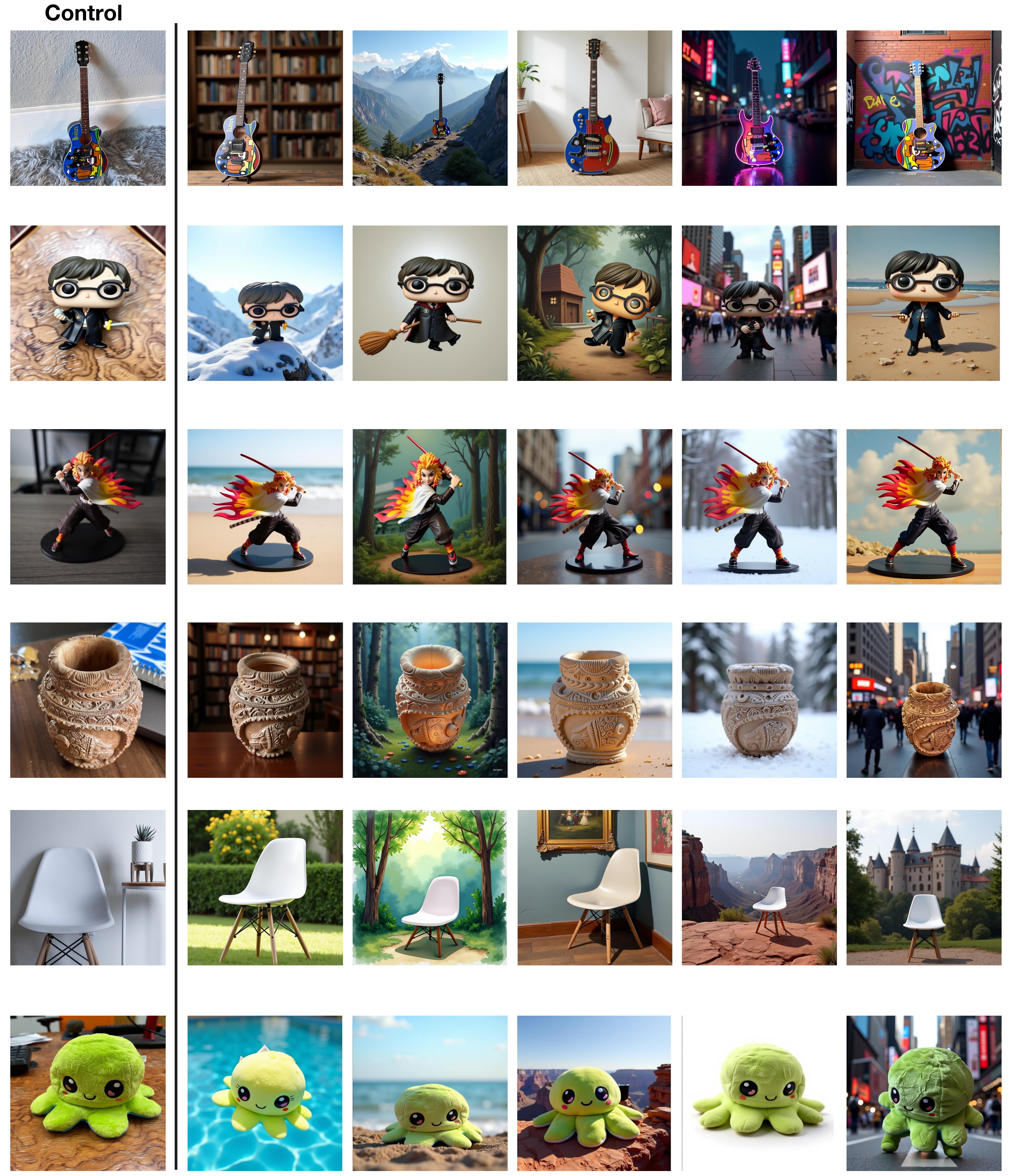}
    \caption{Visualization of subject control generation.}
    \label{fig:sup_subject}
\end{figure*}

\begin{figure*}[!b]
    \centering
    \includegraphics[width=1.\linewidth]{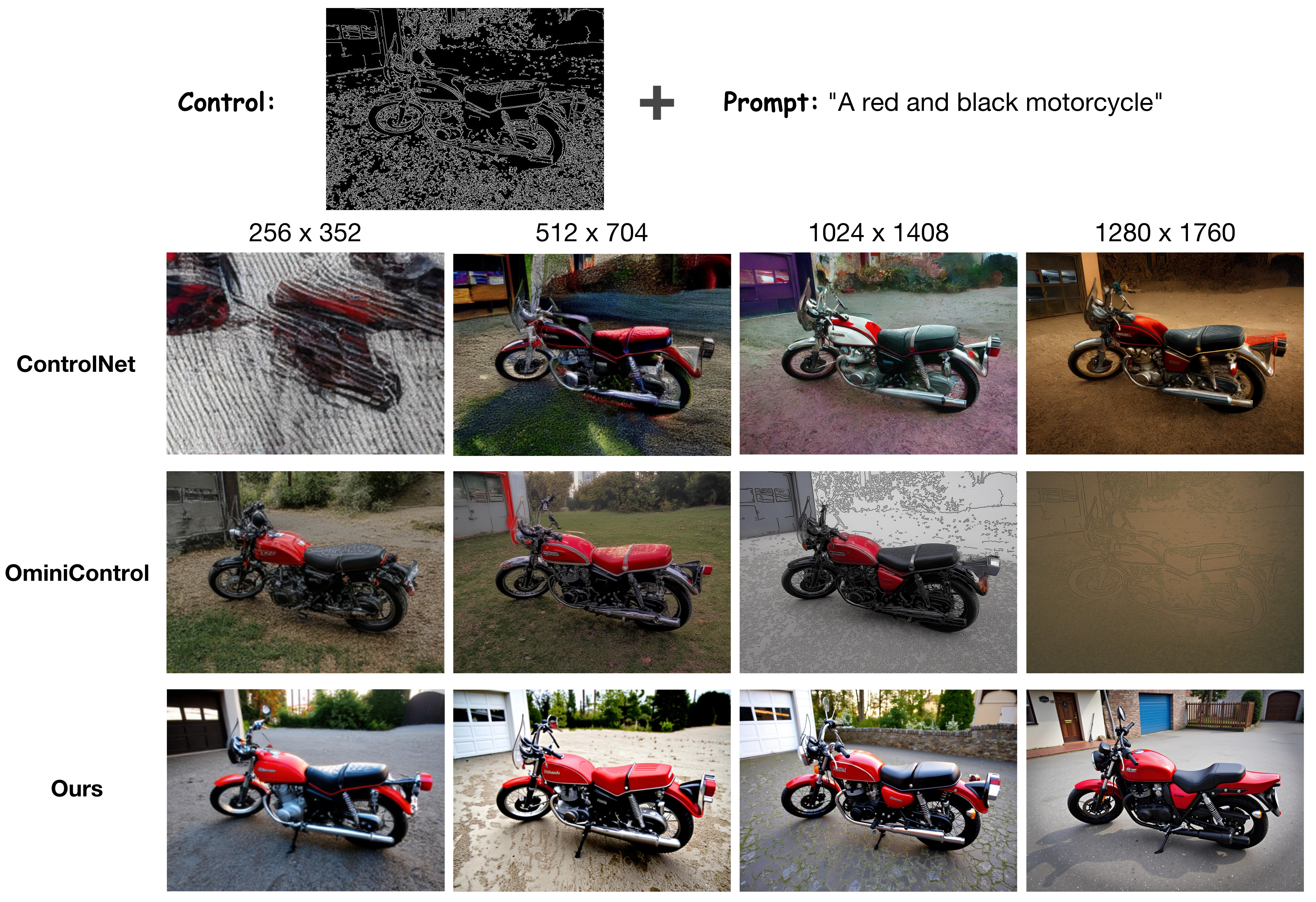}
    \caption{Visual comparison with baseline methods under different resolution generation settings.(zoom in for a better view)}
    \label{fig:sup_res}
\end{figure*}

\clearpage
{
\small

\bibliographystyle{IEEEtran}
\bibliography{reference}
}


\end{document}